\documentclass{article}

\usepackage{microtype}
\usepackage{graphicx}
\usepackage{subcaption}
\usepackage{booktabs} 

\usepackage{hyperref}

\usepackage[preprint]{icml2026}

\usepackage{amsmath}
\usepackage{amssymb}
\usepackage{mathtools}
\usepackage{amsthm}

\newcommand{\ours}{\textbf{RSGround-R1}\xspace}
\usepackage{multirow}
\usepackage[table]{xcolor}
\usepackage{xcolor}
\usepackage{amssymb}  
\usepackage{booktabs}  
\usepackage{colortbl}  
\newcommand{\cmark}{\checkmark}

\newcommand{\xmarkg}{\textcolor{lightgray}{\texttimes}}
\usepackage{bm}

\usepackage{tcolorbox}
\tcbuselibrary{listings, skins, breakable}

\usepackage{xspace}

\usepackage{hyperref}
\usepackage[capitalize,noabbrev]{cleveref}

\def\papertitle{RSGround-R1: Rethinking Remote Sensing Visual Grounding \\ through Spatial Reasoning}

\icmltitlerunning{RSGround-R1: Rethinking Remote Sensing Visual Grounding through Spatial Reasoning}

\begin{document}

\twocolumn[
  \icmltitle{\papertitle}


  \icmlsetsymbol{equal}{*}

  \begin{icmlauthorlist}
    \icmlauthor{Shiqi Huang}{111}
    \icmlauthor{Shuting He}{222}
    \icmlauthor{Bihan Wen}{111}
  \end{icmlauthorlist}

  \icmlaffiliation{111}{School of Electrical and Electronic Engineering, Nanyang Technological University}
  \icmlaffiliation{222}{MoE Key Laboratory of Interdisciplinary Research of Computation and Economics,
Shanghai University of Finance and Economics}

  \icmlcorrespondingauthor{Bihan Wen}{bihan.wen@ntu.edu.sg}

  \icmlkeywords{Machine Learning, ICML}

  \vskip 0.3in
]

\printAffiliationsAndNotice{}

\begin{abstract}
Remote Sensing Visual Grounding (RSVG) aims to localize target objects in large-scale aerial imagery based on natural language descriptions. Owing to the vast spatial scale and high semantic ambiguity of remote sensing scenes, these descriptions often rely heavily on positional cues, posing unique challenges for Multimodal Large Language Models (MLLMs) in spatial reasoning. To leverage this unique feature, we propose a reasoning-guided, position-aware post-training framework, dubbed \ours, to progressively enhance spatial understanding. Specifically, we first introduce Chain-of-Thought Supervised Fine-Tuning (CoT-SFT) using synthetically generated RSVG reasoning data to establish explicit position awareness. Reinforcement Fine-Tuning (RFT) is then applied, augmented by our newly designed positional reward that provides continuous and distance-aware guidance toward accurate localization. Moreover, to mitigate incoherent localization behaviors across rollouts, we introduce a spatial consistency guided optimization scheme that dynamically adjusts policy updates based on their spatial coherence, ensuring stable and robust convergence. Extensive experiments on RSVG benchmarks demonstrate superior performance and generalization of our model.
\vspace{-4mm}
\end{abstract}    
\section{Introduction}
\label{sec:intro}

Remote Sensing Visual Grounding (RSVG) is a fundamental task in geospatial analysis that aims to accurately localize target objects within large-scale aerial imagery based on natural language descriptions~\cite{sun2022visual, zhan2023rsvg, kuckreja2024geochat, li2024vrsbench}. 
This capability is essential for a wide range of applications, including military target identification, urban infrastructure mapping, and large-scale environmental monitoring~\cite{zhan2023rsvg,lobry2020rsvqa,ning2021semantics,lu2017exploring}. 
As remote sensing data from satellites and aerial platforms continue to grow exponentially, RSVG has become an indispensable tool for bridging human language queries with precise geospatial locations.

\begin{figure}[t]
  \centering
   \includegraphics[width=0.95\linewidth]{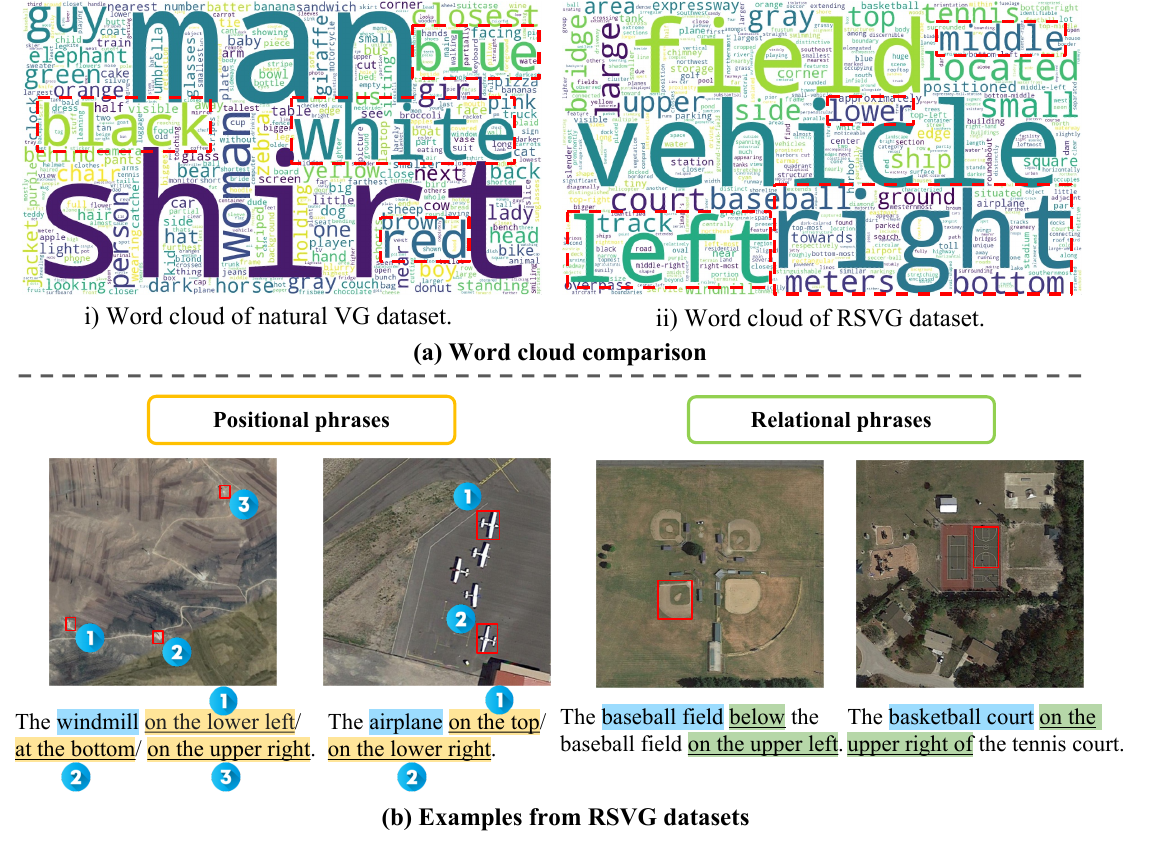}
   \vspace{-2mm}
   \caption{\textbf{(a)} Word cloud comparison between natural VG datasets RefCOCO+ \cite{yu2016modeling}, which relies exclusively on semantic attributes, and RSVG datasets \cite{zhan2023rsvg,liu2024rotated,li2024vrsbench}. Phrase size corresponds to its frequency. \textbf{(b)} Examples from DIOR-RSVG dataset \cite{zhan2023rsvg} with positional and relational phrases.} 
   \label{fig:teaser}
   \vspace{-5mm}
\end{figure}


\begin{figure*}[t]
  \centering
   \includegraphics[width=0.95\textwidth]{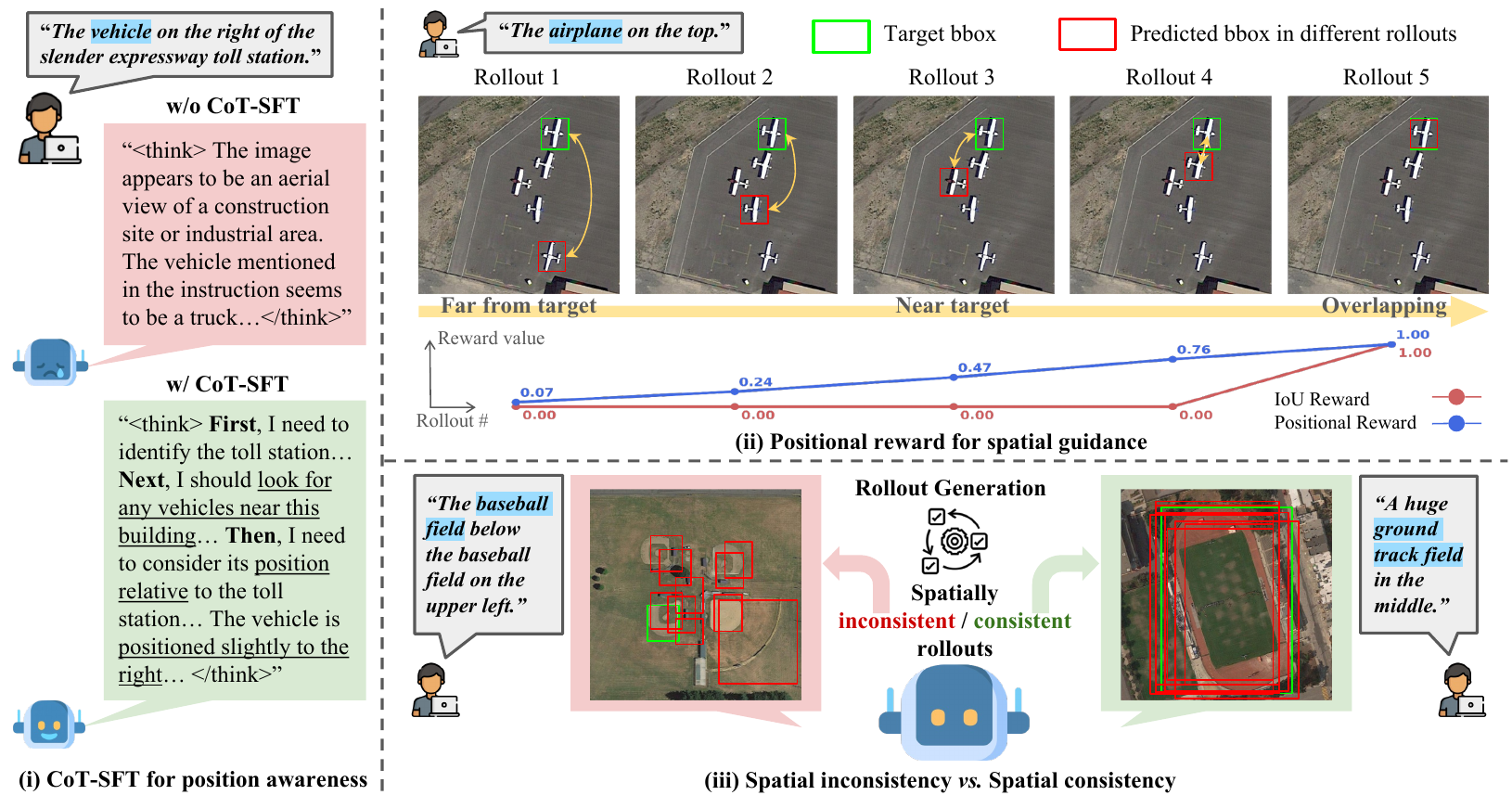}
   \vspace{-2mm}
   \caption{\textbf{(i) CoT-SFT for position awareness.} With CoT-SFT, the model can provide thinking process with clearer, position-aware spatial reasoning. \textbf{(ii) Positional reward for spatial guidance.} Positional reward increases as prediction approaches the target, whereas IoU reward remains zero. \textbf{(iii) Illustration of spatial inconsistency.} Samples representing spatially inconsistent/consistent rollouts.}
   \label{fig:teaser2}
   \vspace{-5mm}
\end{figure*}

Despite recent progress in Vision–Language Models (VLMs) for Earth Observation (EO) tasks~\cite{hu2025rsgpt, liu2024remoteclip, wang2024skyscript, li2024vrsbench}, adapting these models to RSVG remains challenging. 
Remote sensing imagery differs fundamentally from natural images due to its vast spatial scale and high semantic ambiguity inherent in complex scenes~\cite{huang2025score}. 
As shown in Figure~\ref{fig:teaser}(a), natural visual grounding (VG) benchmarks such as RefCOCO+ \cite{yu2016modeling} rely more on semantic attributes, such as color and texture, where target objects are typically salient and distinct.
In contrast, linguistic expressions in RSVG depend predominantly on positional and relational cues. 
As illustrated in Figure~\ref{fig:teaser}(b), target objects in RSVG are often small, ambiguous, and non-unique~\cite{zhan2023rsvg, sun2022visual}, requiring differentiation through positional descriptions such as ``on the lower left'' or ``to the right of the tennis court''. 
This paradigm shift makes spatial reasoning, rather than semantic recognition, the core challenge in RSVG, demanding fine-grained positional awareness across large and complex remote sensing scenes.



 While Multimodal Large Language Models (MLLMs) demonstrate strong generalization abilities across diverse tasks, 
 prior studies have shown that they often struggle with spatial reasoning~\cite{fu2024blink, majumdar2024openeqa, zhang2024countercurate, wang2024picture}. 
 Our preliminary experiment further reveal that MLLMs, such as Qwen2.5-VL~\cite{bai2025qwen2}, fail to produce explicit spatial reasoning without structured spatial cues, as illustrated in Figure \ref{fig:teaser2} (i), resulting in vague or semantically biased outputs for RSVG.
 To overcome this limitation, we introduce a specialized Supervised Fine-Tuning (SFT) stage termed \textbf{Chain-of-Thought SFT (CoT-SFT) for Position Awareness}.
 This stage leverages synthetically generated CoT data specifically designed to instill explicit and interpretable positional reasoning. 
 Through this targeted preparation, the model is primed with a structured understanding of common spatial expressions in RSVG, establishing a solid foundation for subsequent reasoning enhancement.

After establishing positional reasoning awareness through CoT-SFT, we further enhance spatial precision via Reinforcement Fine-Tuning (RFT) with our spatially guided Group Relative Policy Optimization (GRPO) \cite{shao2024deepseekmath}. 
Existing GRPO paradigms \cite{yang2025r1, huang2025vision, shen2025vlm} typically employ Intersection over Union (IoU) as a reward to supervise overall grounding accuracy.
However, as illustrated in Figure \ref{fig:teaser2}(ii), this feedback collapses to zero when the predicted region does not overlap with the ground truth, providing no informative gradient to distinguish near-miss predictions from entirely incorrect ones, e.g., those far from the target.
Consequently, the model receives weak guidance for progressively refining its localization toward the target. 
To overcome this limitation and bridge the gap between high-level reasoning and fine-grained spatial precision in RSVG’s low-cue environment, we introduce a \textbf{Positional Reward for Spatial Guidance}. 
This continuous and distance-sensitive reward formulation provides smooth and informative feedback, effectively encouraging the model to iteratively adjust its predictions closer to the target region during RFT post-training.


Another key challenge we observe in RSVG during RFT training is the spatial inconsistency of rollouts generated from the same query, as demonstrated in Figure \ref{fig:teaser2} (iii).
This inconsistency arises from the characteristics of remote sensing imagery, which typically covers vast areas containing densely distributed and visually similar objects, making it difficult for the model to maintain coherent localization behavior across rollouts.
As a result, the policy model often produces scattered predictions, weakening its spatial reasoning capability and overall grounding stability.
To address this issue, we design a \textbf{Spatial Consistency Guided Optimization Scheme} that explicitly regulates the learning dynamics based on spatial coherence among rollouts. 
The mechanism quantifies the spatial dispersion of predictions within each query group and adaptively assigns higher weights to samples exhibiting greater inconsistency.
By directing the model’s focus toward these spatially diverse cases, the optimization promotes stable, contextually aligned grounding behavior.

In summary, this work presents the first principled study of spatial reasoning in MLLMs for RSVG, from the perspectives as follows:
\vspace{-8pt}
\begin{itemize}
\setlength{\itemsep}{0pt}
    \item We introduce a cold-start CoT-SFT strategy that instills positional reasoning awareness, providing a structured initialization for RFT, leveraging our curated high-quality RSVG CoT dataset.
    \item We propose a positional reward that serves as a dense relaxation of IoU, enabling informative gradient signals even under zero-overlap conditions and facilitating progressive localization refinement.
    \item We develop a spatial consistency guided optimization scheme that modulates learning dynamics based on spatial coherence cues, enhancing stability and reliability during policy updates.
    \item Together, we propose \ours, which endows reasoning ability while achieving strong effectiveness and generalization across multiple RSVG benchmarks, despite using only a small fraction of the training data.
\end{itemize}
\section{Related Work}
\label{sec:related_work}
\subsection{Remote Sensing Visual Grounding}
Remote sensing visual grounding (RSVG) aims to localize a target in aerial imagery from
a natural-language description, requiring fine-grained spatial understanding. Early progress was established by Sun et al. \cite{sun2022visual}, and extended by DIOR-RSVG \cite{zhan2023rsvg} with broaden categories and scene scale. Due to large spatial scale, complex semantics, and high inter-class similarity of aerial scenes \cite{hanyu2024aerialformer, yao2016semantic, huang2025score, huang2025zori}, the expressions in referring comprehensions rely largely on positional or relational phrases to disambiguate visually similar objects \cite{zhan2023rsvg, yuan2023rrsis}. 

Despite progress, most existing RSVG approaches \cite{sun2022visual, zhan2023rsvg} directly adapt general-purpose grounding or detection frameworks without explicitly modeling positional reasoning. These models often focus on category-level alignment rather than geometric understanding, leading to degraded performance under spatial ambiguity or scenes containing non-unique objects. As a result, there remains a clear need for methods capable of capturing explicit positional reasoning and geospatial consistency tailored to the unique characteristics of remote sensing data.

\subsection{Remote Sensing VLMs}
Advances in VLMs for natural images have driven the development of remote sensing foundation models for Earth Observation (EO). Early works focus on image-level understanding using paired image–text data \cite{liu2024remoteclip, wang2024skyscript, zhang2024rs5m}, while more recent MLLM-based models extend to region-level \cite{kuckreja2024geochat, pang2025vhm, muhtar2024lhrs, li2025lhrs, luo2024sky} and pixel-level analysis \cite{shabbir2025geopixel, ou2025geopix, li2025segearth}. However, most existing remote sensing VLMs and MLLMs still rely heavily on vision–language contrastive pretraining \cite{radford2021learning} and supervised fine-tuning on EO tasks, which constrains their reasoning capability and generalization. Recent attempts to introduce RFT into remote sensing tasks \cite{koksal2025tinyrs, muhtar2025quality, yao2025remotereasoner, jiang2025grasp, fiaz2025geovlm, zhang2025geo, liu2025towards} mostly are direct deployment of RL frameworks, without tailoring them to the characteristics of remote sensing data. Moreover, these approaches do not account for the distinctive spatial complexities inherent in geospatial imagery which is explicitly explored in our work.

Meanwhile, recent research reveals that most MLLMs still struggle with spatial reasoning \cite{fu2024blink, majumdar2024openeqa, zhang2024countercurate, wang2024picture}, prompting the development of general benchmarks that explicitly evaluate this ability \cite{liu2024mmbench, liu2023visual, azzolini2025cosmos, tian2025nuscenes} and design of 3D-oriented spatial-aware models \cite{chen2024spatialvlm, cheng2024spatialrgpt}. 

\begin{figure*}[t]
  \centering
    \includegraphics[width=0.95\textwidth]{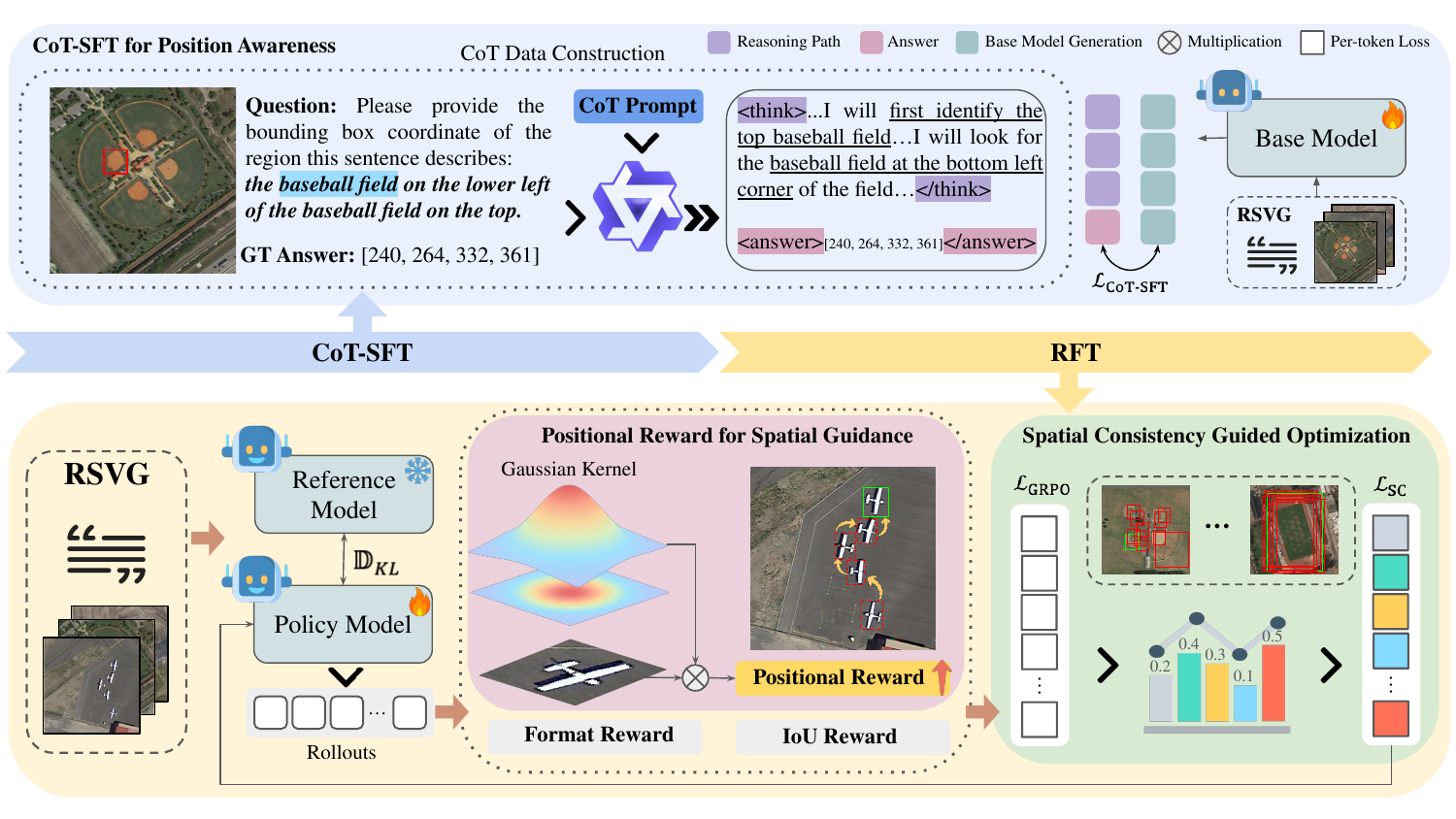}
    \vspace{-2mm}
   \caption{\textbf{Overview of \ours pipeline.} Our framework has two consecutive stages. \textbf{CoT-SFT (Stage 1)}: we curate high-quality, position-aware CoT data to finetune the base model and instill explicit positional understanding. \textbf{RFT (Stage 2)}: after initializing the policy and reference model with the CoT-SFT finetuned based model, we further incentivize spatial reasoning using RFT by GRPO. Along with format and IoU rewards, we introduce a positional reward for progressive localization. To mitigate spatial inconsistent rollouts, we apply a spatial consistency guided optimization that reweights the per token GRPO loss $\mathcal{L}_{\text{GRPO}}$ to yield $\mathcal{L}_{\text{SC}}$, promoting more stable and contextually aligned grounding behavior.}
   \vspace{-4mm}
   \label{fig:framework}
\end{figure*}

\subsection{Reinforcement Fine-Tuning of VLMs}
Reinforcement learning (RL) has become an effective post-training paradigm for enhancing multimodal reasoning in VLMs \cite{huang2025vision, guo2025deepseek, shen2025vlm}. Methods such as DPO and PPO are widely used to align models with human or task-specific feedback \cite{rafailov2023direct, schulman2017proximal, xie2024v, achiam2023gpt, zhai2024fine, ma2024vision}. More recently, GRPO introduces simple rule-based rewards to facilitate structured reasoning and has shown strong generalization on vision-language tasks \cite{shao2024deepseekmath, guo2025deepseek, yang2025r1, huang2025vision, shen2025vlm, bai2025univg}, including referring expression comprehension and open-vocabulary detection \cite{shen2025vlm}. However, most existing RL-based reasoning frameworks remain concentrated on mathematical, coding, or general vision tasks, without specific consideration for remote-sensing scenarios.
\section{Method}
\label{sec:method}

\subsection{Overview}
Our framework follows a two-stage pipeline. \textbf{CoT-SFT (Stage 1)}: we align the base model with high-quality, position-aware CoT data, establishing explicit positional interpretability. \textbf{RFT (Stage 2)}: we further incentivize spatial reasoning through RFT by augmenting GRPO with a positional reward and a spatial consistency guided optimization scheme. Together, these stages enable \ours to learn a deep, non-heuristic understanding of spatial relationships, yielding robust and reliable visual grounding capability for remote-sensing imagery.

\subsection{CoT-SFT for Position Awareness}
The first stage of our framework focuses on priming the model with explicit positional understanding before RFT. While general MLLMs such as Qwen2.5-VL \cite{bai2025qwen2} or InternVL-2.5 \cite{chen2024expanding} possess strong multimodal alignment, they inherently lack spatial interpretability when handling remote sensing imagery.
Their responses to positional queries are typically direct coordinate guesses without intermediate reasoning based on our preliminary inspection. To address this, we introduce the CoT-SFT stage which primes the model with positional understanding through explicit step-by-step reasoning traces.

\paragraph{CoT Data Construction.}
We first generate a CoT dataset focused on positional interpretation for RSVG. Given an annotated image–sentence pair $(I, q, B_{gt})$, where $q$ is a referring expression and $B_{gt}$ is its corresponding ground truth (GT) bounding box, we use the advanced MLLM Qwen2.5-VL-72B \cite{bai2025qwen2} to generate CoT reasoning process. Specifically, as illustrated in Figure \ref{fig:framework}, we provide the model with the question, GT bounding box coordinates, and a predefined CoT prompt, prompting it to generate reasoning processes in the format: \textit{“$<$think$>$thinking process here$<$/think$>$$<$answer$>$GT coordinates$<$/answer$>$”}. 

Each generated sample is denoted as $(I, q, r, B_{gt})$, where $r$ is the step-by-step reasoning path. To ensure data quality, we filter out responses that contain incomplete reasoning chains, inconsistent coordinate formats, or wrong bounding boxes. This process yields in a total of 30k synthetic CoT samples emphasizing spatial reasoning. More details are provided in the Appendices.

\paragraph{CoT-SFT Cold Start Initialization.}
We perform CoT-SFT on the base model Qwen2.5-VL-3B to guide the model in producing explicit and meaningful positional reasoning chains. The training objective is to maximize the conditional likelihood of producing the reasoning and answer sequence $(r, B_{gt})$, given the image-sentence pair $(I, q)$:
\begin{align}
    \mathcal{L}_{\text{CoT-SFT}} = - \mathbb{E}_{(I, q, r, B_{gt}) \sim \mathcal{D}}\left[\sum_{t=1}^{T}\log\pi_{\theta(y_t \mid I, q, y_{<t})}\right],
\end{align}
where $D$ is the CoT dataset, $y$ is the concatenated sequence of $r$ and $B_{gt}$, and $\pi_{\theta}$ represents the model's token distribution. The output model is then serves as the initialization of the policy and reference model in subsequent RFT stage.

This approach primes the model with a structured, step-by-step thinking process for RSVG queries, which gives the model a foundational understanding of the desired behavior before it is asked to explore and learn from reward signals in the more open-ended RL environment.

\subsection{Positional Reward for Spatial Guidance}
While the CoT-SFT stage equips the model with spatial reasoning priors for interpreting positions, it still lacks progressive spatial guidance during RFT. Standard GRPO implementations employ verifiable rewards such as format correctness or task accuracy (e.g., IoU between predicted and GT bounding boxes for VG task). However, IoU provides only a limited spatial signal: any predictions with zero overlap receive zero reward, offering no useful feedback for them to facilitate localization refinement. In the context of RSVG, where targets occupy small portions of extremely large images, this sparsity makes it difficult for the model to improve from completely wrong predictions. To provide continuous and spatially meaningful feedback, we design a positional reward that measures proximity between the predicted and target regions using a Gaussian proximity kernel. This kernel assigns higher values to predictions closer to the ground truth and lower values to distant ones, producing a smooth and progressive reward.

\paragraph{Positional Reward Formulation.}
Let $B_p=\left[x^p_1,y^p_1, x^p_2, y^p_2\right]$ denotes the predicted bounding box and $B_{gt}$ the ground truth. We define a Gaussian kernel $\mathcal{K}(x,y)$ centered at the GT bounding box:
\begin{align}
    \mathcal{K}(x,y)=\exp\!\left(
    -\frac{(x-c_x^{gt})^2}{2\sigma_x^2}
    -\frac{(y-c_y^{gt})^2}{2\sigma_y^2}
    \right), 
\end{align}
where $(c_x^{gt}, c_y^{gt})$ is the ground-truth center, $ \sigma_x=\tfrac{\alpha W}{2}$,  $\sigma_y=\tfrac{\alpha H}{2}$, with $W$ and $H$ being the box width and height. Here, $\alpha$ controls the kernel deviations $\sigma_x$ and $\sigma_y$ based on $W$ and $H$. We apply max-normalization to the kernel:
\begin{align}
    \widehat{\mathcal{K}}(x,y)=\frac{\mathcal{K}(x,y)}{\text{max}_{(u, v)}\mathcal{K}(u,v)}, 
\end{align}

and compute the positional reward as the mean activation of this kernel within the predicted region:
\begin{align}
R_{\text{pos}}(B_p,B_{gt})=\frac{1}{|B_p|}
\sum_{(x,y)\in \mathcal{P}(B_p)}\widehat{\mathcal{K}}(x,y),
\end{align}
where $\mathcal{P}(B_p)$ denotes the set of discrete pixel coordinates within the predicted bounding box. Intuitively, predictions that overlap or lie near to the ground-truth region receive higher scores, thereby encouraging far-off predictions to move toward the target by earning higher reward.

\paragraph{Integration into GRPO}
We combine the positional reward with the standard format reward and IoU reward \cite{shen2025vlm, yang2025r1} to form the overall verifiable signal:
\begin{align}
R=R_{\text{format}}+R_{\text{IoU}}+\beta R_{\text{pos}},
\end{align}
where $\beta$ controls the contribution of positional proximity, balancing local spatial refinement with task correctness. During training, the reward is computed for each rollout and normalized across the group before computing the relative advantage in GRPO.
This formulation allows the model to receive meaningful spatial gradients even when $R_{\text{IoU}}=0$, effectively guiding the predicted boxes to progressively move closer to the target region through continuous positional feedback.

\subsection{Spatial Consistency Guided Optimization}
Due to the intrinsic characteristics of remote sensing imagery such as large spatial scales, visually similar or closely spaced objects within a scene, the rollouts generated frequently display noticeable spatial inconsistency for RSVG. This phenomenon reflects the model’s tendency to produce predictions that vary across different samples of the same query, leading to unstable localization behavior. To mitigate this issue, we introduce a spatial consistency guided optimization scheme that explicitly integrates spatial coherence into the policy update process.

To quantify the spatial consistency among multiple rollouts of the same query, we compute two statistics over their IoU rewards. Given a set of $G$ rollouts, the mean and variance of IoU rewards are defined as:
\begin{align}
\text{m(IoU)}=\frac{1}{G}\sum^G_{g=1}R^{(g)}_{\text{IoU}},
\end{align}
\begin{align}
\text{Var(IoU)}=Var(R^{(1)}_{\text{IoU}}, ..., R^{(G)}_{\text{IoU}}).
\end{align}
These metrics jointly describe the spatial alignment of rollouts: higher m(IoU) indicates accurate localization, whereas lower 
Var(IoU) reflects coherent predictions across rollouts. We then derive an adaptive weighting factor that integrates these two signals into the optimization objective:
\begin{align}
w=\text{exp}(\lambda_{m}(1-\text{m(IoU)})+\lambda_v\text{Var(IoU)}),
\end{align}
which scales the per-token GRPO loss:
\begin{align}
\mathcal{L}_{\text{SC}}=w\cdot \mathcal{L}_{\text{GRPO}}.
\end{align}
Here, $\lambda_m$ and $\lambda_v$ are coefficients controlling the contribution of mean accuracy and spatial variance, respectively. To ensure stability, $w$ is clipped within 
$[w_{\text{min}}, w_{\text{max}}]$, where $w_{\text{min}}=0.5$ and $w_{\text{max}}=3.0$ in our implementation.

\setlength{\tabcolsep}{6.4pt}
\begin{table*}[t] 
    \footnotesize
    \begin{center}
    \caption{Comparison with SOTA models. We evaluate both open-source general-purpose VLMs and RS-specific VLMs under zero-shot inference on RSVG datasets. Additionally, we report results of finetuned RS-VLMs to provide a comprehensive in-domain evaluation. Methods marked with \textsuperscript{$\star$} indicates SFT for 1 epoch, while \textsuperscript{$\dag$} denotes GRPO-based RFT training for only 0.4 epoch using 40\% of the training data. \textbf{Bold} and \underline{underline} represent the best and second best results. We report the average results over three independent runs.} 
    \begin{tabular}{lcccccccccccccc}
    \toprule
    \multirow{2}{*}{\textbf{Method}}
    & \multirow{2}{*}{\textbf{Params}}
    &  \multicolumn{3}{c}{\textbf{DIOR-RSVG}} 
    & \multicolumn{3}{c}{\textbf{VRSBench-VG}} 
    & \multicolumn{3}{c}{\textbf{RRSIS-D}} 
    \\
    
    \cmidrule(lr){3-5} \cmidrule(lr){6-8} \cmidrule(lr){9-11} 
    
    & & \textbf{@0.5} & \textbf{@0.7} & \textbf{mIoU}
    & \textbf{@0.5} & \textbf{@0.7} & \textbf{mIoU}
    & \textbf{@0.5} & \textbf{@0.7} & \textbf{mIoU}
    
    \\
    \midrule 
    
    \multicolumn{7}{l}{\textcolor{gray}{\textit{General Vision-Language Models}}} \\
    MiniGPT-v2 \cite{chen2023minigpt}  &7B & 30.25 & 11.46 & 30.08  & 35.84 & 16.82 & 34.72 & 37.26 & 15.88 & 38.65  \\
    LLaVA-1.5 \cite{liu2024improved} &7B & 16.07 & 4.15 & 23.15 & 15.29 & 3.43 & 23.93 & 16.09 & 4.08 & 24.60  \\
    Qwen2.5-VL-3B \cite{bai2025qwen2}&3B & 37.50 & 26.25 & 37.31 & 38.25 & 19.65 & 37.03 & 38.75 & 26.05 & 39.22  \\
    Qwen2.5-VL-7B \cite{bai2025qwen2}&7B & 39.25 & 25.20 & 38.66 & 43.60 & 22.90 & 41.13 & 41.95 & 26.30 & 41.08  \\
    \cmidrule(r){1-11}
    
     \multicolumn{7}{l}{\textcolor{gray}{\textit{Remote Sensing Vision-Language Models}}} \\
    GeoChat \cite{kuckreja2024geochat}&7B & 25.95 & 4.53 & 29.87 & 14.19 & 2.29 & 23.66 & 24.36 & 3.94 & 30.30  \\
    VHM \cite{pang2025vhm} &7B &  52.15 & 25.63 & 46.01 & 23.48 & 4.43 & 28.92 & 61.39 & 42.92 & 53.10 \\
    SkySenseGPT \cite{luo2024sky}&7B & 22.86 & 5.52 & 27.76 & 11.00 & 2.22 & 21.25 & 24.81 & 6.48 & 29.63\\
    LHRS-Bot \cite{muhtar2024lhrs}&7B & 12.49 & 3.02 & 17.91 & 1.40 & 0.20 & 8.23 & 16.81 & 3.43 & 21.59\\
    LHRS-Bot-Nova \cite{li2025lhrs}&8B & 32.87 & 12.37 & 31.61 & 18.30 & 4.30 & 25.98 & 39.41 & 16.86 & 36.39\\
    \midrule 
    \multicolumn{11}{c}{\textbf{\textit{w/ training}}} \\
    
    \midrule
     \multicolumn{7}{l}{\textcolor{gray}{\textit{Remote Sensing Vision-Language Models}}} \\
    GeoChat w/ SFT\textsuperscript{$\star$} &7B & 46.48 & 13.30 & 43.05 & 59.27 & 32.61 & 52.03 & 50.82 & 20.77 & 45.33  \\
    VHM w/ SFT\textsuperscript{$\star$}  &7B & 61.67 &45.73 & 54.85 & 56.48 & 30.07 & 49.71 & 62.31 & 44.67 & 54.19 \\
    
    \cmidrule(r){1-11} 
    
    Qwen2.5-VL-3B w/ SFT\textsuperscript{$\star$}  &3B & 62.60 & 47.60 & 55.71 & \underline{61.85} & \underline{36.55} & \underline{53.22} & \underline{62.60} & \underline{47.40} & \underline{55.95} \\
    Qwen2.5-VL-3B w/ GRPO\textsuperscript{$\dag$}  &3B & \underline{66.65} & \underline{56.15} & \underline{60.18} & 59.25 & 36.00 & 51.23 & 58.50 & 44.20 & 53.60 \\
    \rowcolor{cyan!10}
     \textbf{RSGround-R1}\textsuperscript{$\dag$} & 3B & \textbf{71.81}& \textbf{58.71} & \textbf{63.38}& \textbf{63.72} & \textbf{38.32}& \textbf{54.68}& \textbf{65.73}& \textbf{50.04} & \textbf{58.35}\\
    \bottomrule
\end{tabular}
\label{tab:sota}
\end{center}
\vspace{-2mm}
\end{table*}

By incorporating this adaptive weighting based on spatial coherence, we derive the resulting spatial consistency loss $\mathcal{L}_{\text{SC}}$. This optimization dynamically emphasizes predictions with higher dispersion, guiding the policy toward stable and spatially consistent grounding behavior.
\section{Experiment}
\label{sec:experiment}

\subsection{Datasets and Metrics}
\paragraph{Training Dataset.}
We train the model on multiple remote sensing visual grounding datasets, including widely used benchmarks DIOR-RSVG \cite{zhan2023rsvg}, as well as datasets with grounding annotations such as VRSBench \cite{li2024vrsbench} and RRSIS-D \cite{liu2024rotated}. We denote the visual grounding subset of VRSBench as VRSBench-VG. Together, these datasets represent a diverse cover diverse object categories, instance densities, and scene complexities, and include a wide range of referring expressions, such as positional, directional, and relational descriptions.

\paragraph{Evaluation Dataset.}
We validate our method from two complementary perspectives: in-domain evaluation and out-of-domain generalization. In-domain evaluation is conducted on the official test splits of DIOR-RSVG, VRSBench-VG, and RRSIS-D. To assess out-of-domain generalization, we further employ the visual-grounding-annotated FAST-T and SOTA-T datasets \cite{wang2023samrs} from GeoPixInstruct \cite{ou2025geopix}, ensuring no overlap with the training data to rigorously assess cross-domain generalization. For models fine-tuned on VRSBench-VG, SOTA-T is excluded since it is derived from DOTA \cite{ding2021object}, which is included in VRSBench-VG.

\paragraph{Evaluation Metrics.}
We adopt the conventional evaluation protocol used in visual grounding. Specifically, we report Accuracy (Acc), where a prediction is regarded as correct if its Intersection over Union (IoU) with the ground-truth bounding box exceeds a predefined threshold (0.5 for Acc@0.5 and 0.7 for Acc@0.7). We also report the mean IoU (mIoU) to provide a more detailed assessment of localization precision and overall grounding performance.

\subsection{Model and Implementation Details}
We employ Qwen2.5-VL-3B-Instruct \cite{bai2025qwen2} as the base model. Our implementation is built upon VLM-R1 \cite{shen2025vlm}, following its default hyperparameter settings unless otherwise specified. For GRPO-based RFT, we train the model for 0.4 epoch using only 40\% of each training dataset, which substantially improves data-efficiency while still achieving superior performance. For experiments with SFT, we finetune each method for 1 epoch using the full training set of each RSVG dataset. In our experiments, the positional reward is configured with $\beta=0.1$ and $\alpha=2.5$. For spatial consistency guided optimization, we set $\lambda_m=1.2$ and $\lambda_v=1$.

\setlength{\tabcolsep}{2.2pt}
\begin{table*}[t]
\footnotesize
  \begin{center}
  \caption{Out-of-domain generalization. We assess the models’ generalization capability on the FAST-T and SOTA-T datasets, ensuring that no data overlap exists between the training and testing sets. \textbf{Bold} and \underline{underline} represent the best and second best results.}
  \begin{tabular}{lccccccccccccccc}
  \toprule
    & \multicolumn{6}{c}{\textbf{DIOR-RSVG}} & \multicolumn{3}{c}{\textbf{VRSBench-VG}} & \multicolumn{6}{c}{\textbf{RRSIS-D}}\\
  \cmidrule(r){2-7} \cmidrule(r){8-10} \cmidrule(r){11-16}
   \textbf{Method} & \multicolumn{3}{c}{\textbf{FAST-T}} & \multicolumn{3}{c}{\textbf{SOTA-T}} &  \multicolumn{3}{c}{\textbf{FAST-T}}  & \multicolumn{3}{c}{\textbf{FAST-T}} & \multicolumn{3}{c}{\textbf{SOTA-T}} \\
  \cmidrule(r){2-4} \cmidrule(r){5-7} \cmidrule(r){8-10}
  \cmidrule(r){11-13} \cmidrule(r){14-16}
    & \textbf{@0.5} & \textbf{@0.7} & \textbf{mIoU}
    & \textbf{@0.5} & \textbf{@0.7} & \textbf{mIoU}
    & \textbf{@0.5} & \textbf{@0.7} & \textbf{mIoU}
    & \textbf{@0.5} & \textbf{@0.7} & \textbf{mIoU}
    & \textbf{@0.5} & \textbf{@0.7} & \textbf{mIoU}\\
    \cmidrule(r){2-16}
GeoChat w/ SFT\textsuperscript{$\star$} & 20.81&4.65 &25.77 & 15.87& 3.99&20.66 &26.93 &8.17 &28.21 &11.73 & 2.21&18.56 &9.11 &2.60 &14.14 \\
VHM w/ SFT\textsuperscript{$\star$} & 16.69&4.59 &21.88 & 14.40& 4.56&18.43 &29.08 &11.71 &28.38 &9.35 & 2.09& 15.75 & 9.11 & 2.60 &13.24 \\
Qwen2.5-VL-3B w/ SFT\textsuperscript{$\star$} &27.78 &10.92 &29.80 &27.58 &9.11 &27.78 &43.70 &21.10 &37.02 &27.55 &11.11 &29.65&27.99 & 9.36 &27.63\\
Qwen2.5-VL-3B w/ GRPO\textsuperscript{$\dag$} &\underline{36.55} &\underline{18.44} &\underline{35.66} &\underline{34.58} &\underline{17.58} &\underline{31.27} &\underline{47.49} &\underline{27.36} &\underline{39.63} &\underline{33.49} &\textbf{17.19} &\textbf{32.92} &\underline{30.59} &\underline{15.20} &\underline{28.20} \\
\rowcolor{cyan!10}
\textbf{RSGround-R1}\textsuperscript{$\dag$} &\textbf{38.59} &\textbf{19.21} &\textbf{35.96} &\textbf{38.00} &\textbf{19.69} &\textbf{33.25} &\textbf{49.71} &\textbf{28.05} &\textbf{39.93} &\textbf{35.49} &\underline{16.7} &\underline{32.45} &\textbf{32.79} &\textbf{15.54} &\textbf{29.50} \\
    \toprule
  \end{tabular}
\label{tab:ood}
\vspace{-7mm}
\end{center}
\end{table*}

\subsection{Main Results}
\paragraph{Comparison with SOTA models}
Table \ref{tab:sota} compares our method with both general-purpose and remote-sensing-specific VLMs. Among general VLMs, Qwen2.5-VL \cite{bai2025qwen2} consistently outperforms MiniGPT-v2 \cite{chen2023minigpt} and LLaVA-1.5 \cite{liu2024improved}, and even outperforms several RS-specific models such as GeoChat \cite{kuckreja2024geochat}, SkySenseGPT \cite{luo2024sky}, LHRS-Bot \cite {muhtar2024lhrs} and LHRS-Bot-Nova \cite{li2025lhrs}, highlighting the limitations of existing RS VLMs in addressing remote-sensing visual grounding. While RS-specific models improve after fine-tuning and GRPO-based RFT shows competitive performance, our method consistently achieves the best results. Notably, \ours surpasses direct GRPO by over 5\% in Acc@0.5 using only 0.4 epoch (40\% data), demonstrating superior learning effectiveness and generalization.

\vspace{-3mm}
\paragraph{Out-of-domain Generalization}
Table \ref{tab:ood} evaluates the out-of-domain generalization on the FAST-T and SOTA-T benchmarks, which introduce significant regional and sensor shifts. SFT-based methods perform worse than GRPO-based counterparts, highlighting the difficulty of transferring grounding ability through supervised tuning alone. \ours consistently outperforms the GRPO baseline, especially on SOTA-T, demonstrating strong robustness and transferable spatial priors under substantial domain shifts.

\begin{figure*}[t]
  \centering
    \includegraphics[width=0.95\textwidth]{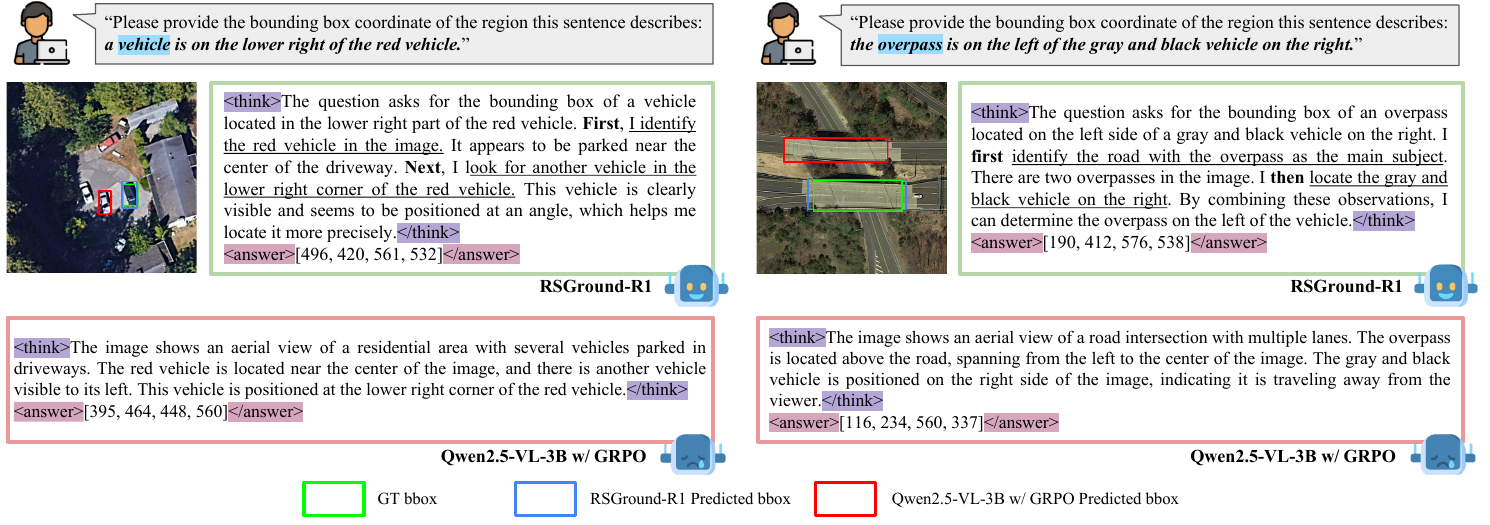}
    \vspace{-2mm}
   \caption{Qualitative comparison. \ours produces more plausible reasoning trajectories and more accurate bounding box predictions compared to the baseline Qwen2.5-VL-3B w/ GRPO. More visualizations can be found in the Appendices.}
   \vspace{-6mm}
   \label{fig:qualitative}
\end{figure*}

\subsection{Ablation Study}
\paragraph{Component Analysis}
\begin{table}[t]
\setlength\tabcolsep{8.4pt}
\footnotesize
  \begin{center}
  \caption{Component Analysis: CoT-SFT, positional reward $R_\text{pos}$ and spatial consistency guided optimization $\mathcal{L}_{\text{SC}}$.}
  \begin{tabular}{cccccc}
  \toprule
  \multicolumn{3}{c}{\textbf{Components}} & \multicolumn{3}{c}{\textbf{DIOR-RSVG}} \\
   \cmidrule(r){1-3}   \cmidrule(r){4-6}    
   \textbf{CoT-SFT} &\boldsymbol{$R_{\textbf{pos}}$}&  \boldsymbol{$\mathcal{L}_\textbf{SC}$}
 & \textbf{@0.5} & \textbf{@0.7} & \textbf{mIoU}  \\
  \toprule
 \xmarkg& \xmarkg& \xmarkg&  66.65  &  56.15  & 60.18 \\
 \cmark& \xmarkg& \xmarkg &  68.45  &  56.70  &  60.43\\
 \xmarkg& \cmark& \xmarkg &  68.10  &  56.85  &  61.12 \\
  \xmarkg& \xmarkg& \cmark &  69.95  &  57.65  & 62.44\\
  \rowcolor{cyan!10}
 \cmark& \cmark& \cmark &  \textbf{71.81}  &  \textbf{58.71}  &  \textbf{63.38}\\
 
 \arrayrulecolor{black}\bottomrule
  \end{tabular}
\label{tab:ca}
\vspace{-6mm}
\end{center}
\end{table}

Table \ref{tab:ca} summarizes the contribution of each component on the DIOR-RSVG dataset. The first row without any components denotes the pure GRPO-based RFT baseline. As seen from the second row, incorporating CoT-SFT yields about a 3\% improvement in Acc@0.5, validating the importance of establishing a positional reasoning foundation. The additional gain achieved by introducing the positional reward $R_\text{pos}$ also demonstrates the effectiveness in guiding the model toward spatially aligned predictions. Integrating the spatial consistency loss $\mathcal{L}_{\text{SC}}$ provides 3.3\% improvement in Acc@0.5 by stabilizing predictions across rollouts. When all three components are combined, the model achieves the best overall performance, confirming their complementary contributions to robust spatial reasoning and grounding precision.

\paragraph{Positional Reward}
\setlength{\tabcolsep}{11.6pt}
\begin{table}[t] 
    \footnotesize
    \begin{center}
    \caption{Positional reward for spatial guidance: comparisons with alternative distance-aware reward functions.}
    \begin{tabular}{lccc}
    \toprule
    \multirow{2}{*}{\textbf{Reward Functions}}
    &  \multicolumn{3}{c}{\textbf{DIOR-RSVG}} 
    \\
    \cmidrule(lr){2-4}  
    & \textbf{@0.5} & \textbf{@0.7} & \textbf{mIoU}\\
    \cmidrule(lr){1-4}  
    \textbf{IoU} \textcolor{gray}{[Baseline]}& 66.65  &  56.15 & 60.18 \\
    \textbf{+GIoU} & 65.10 & 54.05 & 58.87\\
   \textbf{+DIoU} & 63.60 & 53.45 & 57.43\\
   \textbf{+Center distance} & 67.40 & 56.33 & 60.81\\
    \rowcolor{cyan!10}
    \textbf{+\boldsymbol{$R_{\textbf{pos}}$}}\textcolor{gray}{[Ours]}& \textbf{68.10} & \textbf{56.85} & \textbf{61.12}\\
    \bottomrule
\end{tabular}
\label{tab:pr}
\vspace{-3mm}
\end{center}
\end{table}

Table \ref{tab:pr} compares our positional reward $R_\text{pos}$ with other distance-aware functions such as GIoU, DIoU and center distance as rewards. Although they provide non-zero gradients under non-overlapping conditions, applying them as reward functions brings limited or even negative effects. Unlike box-level or distance penalties that can induce noisy and high-variance signals in policy learning, $R_\text{pos}$ offers smooth, scale-adaptive proximity feedback, leading to more stable and informative optimization.

\vspace{-4mm}
\begin{figure}[t]
  \centering
   \includegraphics[width=0.85\linewidth]{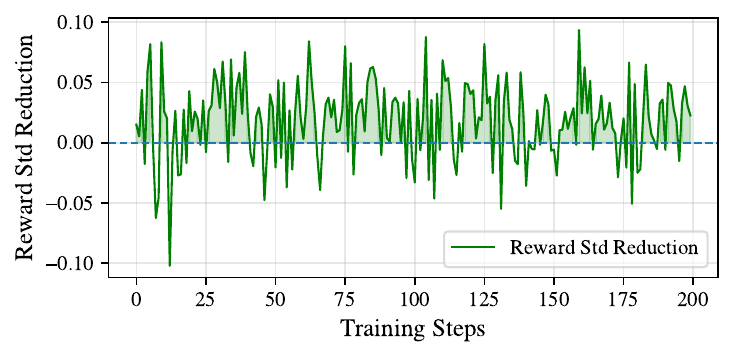}
   \vspace{-2mm}
   \caption{Spatial consistency guided optimization: reward std reduction \textit{vs.} baseline on DIOR-RSVG dataset.}
   \label{fig:std}
   \vspace{-6mm}
\end{figure}

\paragraph{Spatial Consistency}
\begin{table}[t]
\setlength\tabcolsep{10.5pt}
\footnotesize
  \begin{center}
  \caption{Spatial consistency guided optimization: effect of m(IoU) and Var(IoU) in the spatial consistency weighting.}
  \begin{tabular}{ccccc}
  \toprule
   \multirow{2}[6]{*}{\textbf{m(IoU)}} &  \multirow{2}[6]{*}{\textbf{Var(IoU)}} & \multicolumn{3}{c}{\textbf{DIOR-RSVG}} \\
      \cmidrule(r){3-5}    
    &  & \textbf{@0.5} & \textbf{@0.7} & \textbf{mIoU}  \\
  \toprule
 \xmarkg& \xmarkg&  66.65  &  56.15  &  60.18 \\
 \cmark& \xmarkg&  68.68  &  56.55  &  61.11\\
 \xmarkg& \cmark&  67.80  &  56.47  &  60.88\\
 \rowcolor{cyan!10}
 \cmark& \cmark&  \textbf{69.95}  &  \textbf{57.65}  &  \textbf{62.44} \\
 \arrayrulecolor{black}\bottomrule
  \end{tabular}
\label{tab:spacon}
\vspace{-6mm}
\end{center}
\end{table}
Table \ref{tab:spacon} analyzes the effect of incorporating mean and variance terms in the spatial consistency weighting. Using only the mean IoU reward emphasizes overall localization quality, leading to a clear improvement of 2\% in Acc@0.5. Incorporating only the variance term enhances stability by penalizing inconsistent rollouts, resulting in a 1\% gain. Combining both m(IoU) and Var(IoU) achieves the best performance, demonstrating that joint consideration enables more balanced policy updates and improves grounding precision and robustness.

Figure \ref{fig:std} visualizes the reduction in reward standard deviation after applying \boldsymbol{$\mathcal{L}_\textbf{SC}$}. The consistently lower variance across training steps suggests that spatial consistency guided optimization effectively stabilizes rollout behavior, leading to smoother and more reliable policy optimization.

\vspace{-2.5mm}
\subsection{Qualitative Comparison}
The comparison highlights the distinct reasoning and localization behaviors between RSGround-R1 and Qwen2.5-VL-3B w/ GRPO. Qwen2.5-VL-3B w/ GRPO produces a concise yet surface-level response, directly describing visible objects without explicitly decomposing the spatial relationship. Its bounding box prediction aligns with the target object class but shows a positional offset. In contrast, \ours demonstrates a more structured and stepwise reasoning process. This progressive reasoning yields more accurate localization and exemplifies the benefits of Chain-of-Thought SFT combined with spatially guided RFT. Overall, \ours exhibits superior interpretability and spatial awareness, reasoning through multi-step relationships rather than relying on direct visual matching alone.

\vspace{-2mm}
\subsection{Performance under Extended Training}
\begin{figure}[t]
  \centering
   \includegraphics[width=0.85\linewidth]{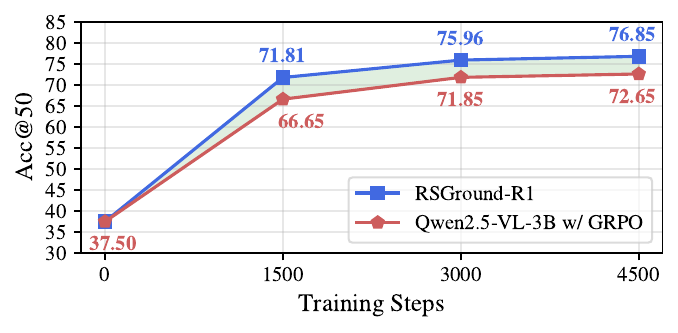}
   \vspace{-2mm}
   \caption{Performance under extended training process beyond the default 0.4-epoch (1500 steps) setting on DIOR-RSVG dataset. More comparisons are provided in the Appendices.}
   \vspace{-6mm}
   \label{fig:step}
\end{figure}

Figure \ref{fig:step} illustrates the sustained benefits of our design by extending training beyond 0.4 epoch (1500 steps) on the DIOR-RSVG dataset and evaluating performance at 1500-step intervals. As training progresses, our method continues to yield steady gains and consistently outperforms the GRPO fine-tuning baseline. These results show that positional reasoning and spatial consistency optimization not only accelerate early-stage learning but also deliver continuous improvements during prolonged training, demonstrating the robustness and scalability of our approach.

\vspace{-3mm}
\section{Conclusion}
\label{sec:conclusion}
\vspace{-1mm}
In this work, we presented \ours, a reasoning-guided, position-aware post-training framework for remote sensing visual grounding that addresses the task’s inherent spatial reasoning challenges. By combining CoT-SFT using our curated CoT data with reinforcement learning enhanced by positional reward and spatial consistency optimization, our approach improves localization accuracy, training stability, and generalization across diverse benchmarks. The interpretability and robustness of our model present a promising direction for advancing geospatial understanding.

\section*{Impact Statement}
While remote sensing visual grounding has beneficial applications in urban planning, disaster assessment, and environmental monitoring, it may also be subject to dual-use risks, including misuse for surveillance or military purposes. We emphasize that the proposed model is intended for research and large-scale scene understanding, rather than individual-level monitoring. Beyond these considerations, we do not identify additional impacts that require specific discussion.

{
    \bibliographystyle{icml2026}
    \bibliography{main}
}

\newpage
\appendix
\onecolumn


\end{document}